\documentclass[5p,times]{elsarticle}

\usepackage{lineno}
\modulolinenumbers[5]
\usepackage{picins}
\usepackage{graphicx}
\usepackage{algorithm}
\usepackage{algorithmic}
\usepackage{amsmath}
\usepackage{amssymb}
\usepackage{mathtools}
\usepackage{centernot}
\usepackage{algorithmic}
\usepackage{multirow}
\usepackage{enumerate}
\usepackage{multicol}
\usepackage{multirow}
\usepackage{booktabs}
\usepackage[justification=centering]{caption}
\usepackage[numbers]{natbib}

\newcommand{\bx}[0]{\mathbf{x}}

\newcommand{\be}[0]{\mathbf{e}}
\newcommand{\bs}[0]{\mathbf{s}}
\newcommand{\bw}[0]{\mathbf{w}}

\newcommand{\bn}[0]{\mathbf{n}}
\newcommand{\bq}[0]{\mathbf{q}}

\newcommand{\bzero}[0]{\mathbf{0}}
\newcommand{\mA}[0]{\mathcal{A}}
\newcommand{\mL}[0]{\mathcal{L}}
\newcommand{\mS}[0]{\mathcal{S}}
\newcommand{\mR}[0]{\mathcal{R}}
\newcommand{\mD}[0]{\mathcal{D}}

\newcommand{\boldeta}[0]{{\boldsymbol \eta}}

\begin{document}
\begin{frontmatter}
\title{Tighter Bound Estimation of Sensitivity Analysis for Incremental and Decremental Data Modification}

\author[add1]{Kaichen~Zhou}
\ead{rui.zhou@ox.cs.ac.uk}
\author[add2]{Shiji~Song\corref{cor1}}
\ead{shijis@mail.tsinghua.edu.cn}
\author[add2]{Gao~Huang}
\ead{gaohuang@tsinghua.edu.cn}
\author[add2]{Wu~Cheng}
\ead{wuc@tsinghua.edu.cn}
\author[add2]{Quan~Zhou}
\ead{zhouq10@mail.tsinghua.edu.cn}
 
\cortext[cor1]{Please address correspondence to Author Shiji~Song}
\address[add1]{Department of Computer Science, University of Oxford}
\address[add2]{Department of Automation, Tsinghua University}

\begin{abstract}
In large-scale classification problems, the data set always be faced with frequent updates when a part of the data is added to or removed from the original data set. In this case, conventional incremental learning, which updates an existing classifier by explicitly modeling the data modification, is more efficient than retraining a new classifier from scratch. However, sometimes, we are more interested in determining whether we should update the classifier or performing some sensitivity analysis tasks. To deal with these such tasks, we propose an algorithm to make rational inferences about the updated linear classifier without exactly updating the classifier. Specifically, the proposed algorithm can be used to estimate the upper and lower bounds of the updated classifier's coefficient matrix with a low computational complexity related to the size of the updated dataset. Both theoretical analysis and experiment results show that the proposed approach is superior to existing methods in terms of tightness of coefficients' bounds and computational complexity.

\end{abstract}
\end{frontmatter}

\section{INTRODUCTION}
Solving the large-scale classification task with datasets in different situations is widely studied in machine learning. e.g., \cite{zhang2016unsupervised} study clustering approach for the data set with missing data; \cite{ristin2016incremental} design the Random Forests for the dynamically growing data set. One of the most common conditions that may be encountered by the large-scale classification task is that part of the data set is modified. However, directly retraining the classifier is time-consuming and will waste a lot of computational resources. The so-called incremental learning is designed for the case in which we want to update the classifier in a different way when the data set is modified \cite{syed1999incremental, cauwenberghs2001incremental, Cho2013}. In the literature, there are mainly two categories of incremental learning methods.

For the first category, the updated classifier can be explicitly derived based on an optimization technique called parametric programming. For example, \cite{gu2015incremental} extend the online $\nu$-support vector classification algorithm to the modified formulation and presents an effective incremental support vector ordinal regression algorithm, which can handle a quadratic formulation with multiple constraints, where each constraint is constituted of equality and inequality. \cite{karasuyama2009multiple} develop an extension of the incremental and decremental algorithm which can simultaneously update multiple data points.

As to the second category, some warm-start approaches without explicitly deriving the updated formulation can also help reduce the incremental learning costs. \cite{tsai2014incremental} find that the warm start setting is in general more effective to improve the primal initial solution than the dual and that the warm-start setting could speed up a high-order optimization method more effectively than a low-order one. \cite{shilton2005incremental} propose a new incremental learning algorithm involving using a warm-start algorithm for the training of support vector machine, which takes advantage of the natural incremental properties of the standard active set approach to linearly constrained optimization problems. 

Conventional incremental learning algorithms aim to solve the primal problem or the dual problem of the optimization problem, which still requires the relatively high cost of computing. \cite{xu2018new} propose an incremental support vector machines based on Markov re-sampling (MR-ISVM) whose computational complexity is up to $O(N^3)$, where $N$ is the number of samples. However, sometimes it is unnecessary to figure out the exact values of the updated classifier's coefficient vector when we only need to estimate the bounds of the updated classifier's coefficient vector. Moreover, most conventional incremental learning algorithms can only be applied to certain classification models, which largely limits their generalizability. For example, \cite{ruping2002incremental} propose an approach for incremental learning specialized for Support Vector Machines. Ren et al. \cite{ren2010incremental} put forward an incremental algorithm only for bidirectional principal component analysis used in pattern recognition and image analysis. For these reasons, the sensitivity analysis with the broad applicability of the updated classifier is necessary.

The aim of sensitivity analysis is not to obtain the exact values of the updated classifier's parameters, but to estimate the bounds of the updated classifier's parameters. It can help researchers understand how uncertainty in the output of a model (numerical or otherwise) can be apportioned to different sources of uncertainty in the model input \cite{saltelli2004sensitivity}. Since sensitivity analysis doesn't require the calculation of exact values of the updated classifier, it can be executed efficiently. Recently, Okumura et al. \cite{okumura2015kdd} proposed a sensitivity analysis framework that can be used to estimate the bounds of a general linear score for the updated classifier.

The performance of sensitivity analysis highly relies on the tightness of bounds it derived. Inspired by recent papers
studying feature screen in the L1 sparse learning framework \cite{ghaoui2010safe,liu2013safe,wang2014safe,xiang2017screening}, we make use of a composite region test and propose a sensitivity analysis framework which can infer bounds for relevant values about the coefficient vector of the updated classifier. Our work aims to improve the algorithm proposed in \cite{okumura2015kdd} and our algorithm can estimate tighter lower and upper bounds of a general linear score in the form of $\boldeta^\top \bw_1^\star$, where $\bw_1^\star$ is the coefficient vector of the updated classifier, and $\boldeta^\top$ is the vector which has the same dimension as $\bw_1^\star$. 

As the proposed algorithm does not have to solve any primal problem or any relevant dual problem of the optimization problem, it largely decreases the computing complexity compared with conventional incremental algorithms and its computing complexity only depends on the number of modified instances, including the number of removed instances and added instances. Besides, the proposed algorithm is more robust than the algorithm proposed in \cite{okumura2015kdd}. Even with a relatively large modification on the data set, the proposed algorithm can still make accurate estimations. And the proposed algorithm only requires that the estimated classifier is calculated based on the differentiable convex L2-regularization loss function, the proposed algorithm can be applied to diversified algorithms.

The estimation about lower and upper bounds of a general linear score in the form of $\boldeta^\top \bw_1^\star$ has numerous applications in a wide range of fields. Firstly, with this scale, the moment when the classifier should be updated can be easily estimated, which can prevent us from making unnecessary updating or omitting necessary updating. Secondly, when the scale is enough tight, the classification results of the updated classifier can be also estimated based on the linear score, which can ensure both low computing complexity and high classification accuracy. Thirdly, the proposed algorithm can also be combined with some model selection algorithms to reduce the computing time of model selection algorithms.

Our contribution can be listed as follow:
\begin{itemize}
    \item Firstly, the proposed sensitivity analysis algorithm can estimate tight bounds of a general linear score about the coefficient vector for the updated classifier.
    \item Secondly, the proposed algorithm has lower computing complexity and higher robustness compared to existing methods.
    \item Thirdly, the proposed algorithm has better applicability than existing methods, as it only requires that the estimated classifier is based on the differentiable convex L2-regularization loss function.
    \item Fourthly, the proposed algorithm has numerous applications that can improve efficiency when solving a large-scale classification problem.
\end{itemize}

The rest of paper is organized as follows. In Section II, we will define some necessary mathematical notations and give a brief overview of the problem setup. Then we will describe two sensitivity analysis tasks that the proposed algorithm can deal with. In Section III, we will give the proof of the proposed algorithm and apply it to the sensitivity analysis tasks proposed in Section II. In Section IV, we will present the detail of our simulation and the analysis of simulation results. In Section V, we will conclude our work and discuss the future direction of our work.

\section{NOTATION AND BACKGROUND}
In this section, we first describe the background of our problem, including the mathematical notation, conventional methods, and our proposed algorithm. Then, we describe two sensitivity analysis tasks that our proposed algorithm can be applied to.

\subsection{Problem Setup}
The problem we study in this paper is that there exists a trained classifier on the original data set, while the current data set is modified by a small number of instances. Instead of retraining a new classifier from scratch with high computing cost, we propose an algorithm to make an inference, in an efficient manner, about the bounds of the updated linear classifier's coefficient vector.

\par We denote scalars in regular ($C$ or $b$), vectors in bold ($\bx$) and matrices in capital bold ($\mathbf{X}$). Specific entries in vectors or matrices follow the corresponding convention, $i.e.$, the $i^{th}$ dimension of vector $\mathbf{x}$ is $x_i$ and $\mathbf 1$ is a column vector where all of its elements are $1$. 

\par We use $\{(\bx_i,y_i)\}_{i \in\mD_0}$ and $\{(\bx_i,y_i)\}_{i \in\mD_1}$ to denote the original data set and the updated training data set respectively, where $\bx_i\!\in\! \mR^d$ and $y_i$ is a binary scalar. The number of training instances in the original training set and the updated training set are denoted as $n_0$ and $n_1$. We consider the scenario where we have an existing classifier $f(\bx;\bw_0)$, then a small amount of instances are added to or removed from the original data set. We denote the set of added and removed instances as $\{(\bx_i,y_i)\}_{i \in \mA}$ and $\{(\bx_i,y_i)\}_{i \in \mS}$, and, $n_{\mA}$ and $n_{\mS}$ are the number of instances in set $\mA$ and $\mS$ respectively. Thus, we will have
\begin{align}
n_{1}-n_{0}= n_{\mA}-n_{\mS}.
\end{align}
We define a new variable
\begin{align}
P_{up}= (n_{\mA}+n_{\mS})/n_{0},
\end{align}
which is used to describe the ratio of the modified data to the original data set.
\par Here we consider a class of L2 regularized linear classification problems with the convex loss function, hence, the original and updated classifiers, trained respectively with the original data set and the modified data set, are defined as
\begin{align}
\bw_0^\star \coloneqq &\arg\min_{\mathbf w}\ \frac{C}{2}\|\mathbf w\|^2+ \frac{1}{n_0}\sum_{i\in \mD_0} \ell(y_i,f(\bx_i;\bw)),  \label{eq:old_svmPrimal}    \\
\bw_1^\star \coloneqq &\arg\min_{\mathbf w}\ \frac{C}{2}\|\mathbf w\|^2+ \frac{1}{n_1}\sum_{i\in \mD_1} \ell(y_i,f(\bx_i;\bw)),  \label{eq:new_svmPrimal}
\end{align}
where $\bw_0^\star$ and $\bw_1^\star$ are the optimal solution to the equations \eqref{eq:old_svmPrimal} and \eqref{eq:new_svmPrimal} respectively; $C\!>\!0$ denotes the regularization constant and $\ell(y_i,f(\bx_i;\bw))$ is a differentiable and convex loss function. When $\ell$ is a squared hinge loss function, i.e.,
\begin{equation}
\ell(y_i,f(\bx_i;\bw)) = \max(1-y_i\mathbf w^\top \bx_i,0)^2,
\label{eq:svmPrimal}
\end{equation}
the problem in \eqref{eq:old_svmPrimal} or \eqref{eq:new_svmPrimal} corresponds to SVM with squared hinge loss (L2-SVM) \cite{steinwart2005consistency, lee2013study} \footnote{Please note that in this paper we do not include any bias term $b$, since we can deal with this term by appending each instance with an additional dimension, i.e., $\bx_i^\top \leftarrow [\bx_i^\top, 1] \quad  \bw^\top \leftarrow [\bw^\top, b]$}. When $\ell$ is a logistic regression loss function, i.e.,
\begin{equation}
\ell(y_i,f(\bx_i;\bw)) = \log(1+\exp (-y_i\mathbf w^\top \bx_i)),
\label{eq:logistic}
\end{equation}
the problem in \eqref{eq:old_svmPrimal} or \eqref{eq:new_svmPrimal} corresponds to L2-regularized logistic regression \cite{zhang2015resting, cox1958regression}. For any $i\in \mD_0\cup \mD_1$, we denote individual loss and the gradient of the individual loss as
\begin{align}
&\ell_i(\bw) = \ell(y_i,f(\bx_i;\bw)),   \\
&\nabla\ell_i(\bw) = \frac{\partial}{\partial\bw}\ell(y_i,f(\bx_i;\bw)).
\end{align}
\par Our main interest in this paper is to avoid calculating $\bw_1^\star$ explicitly as in conventional incremental learning algorithms which has high computing cost when the whole data set is large. Instead, we aim to bound the linear score relevant to the coefficient vector of the updated classifier 
\begin{equation}
L(\boldeta^\top \bw_1^\star) \leq \boldeta^\top \bw_1^\star \leq U(\boldeta^\top \bw_1^\star),
\label{eq:score}
\end{equation}
where $\boldeta$ can be any $d$ dimension vector in $\mR^d$; $L(\boldeta^\top \bw_1^\star)$ and $U(\boldeta^\top \bw_1^\star)$ are the lower and upper bounds of $\boldeta^\top \bw_1^\star$. In addition, our framework can also be applied to nonlinear classification problems with kernel trick, i.e., $\forall i \in \mD_0 \cup \mD_1$, $\boldeta^\top \bw_1^\star$ can be represented by the kernel function. Next, we introduce two applications which we use to verify the performance of the proposed algorithm.

\subsection{Sensitivity analysis tasks}
\subsubsection{Task 1. Sensitivity of coefficients}
\label{sec:sensiCoeff}
When a small amount of instances is added or removed from the original data set, we care about the change about the coefficient vector of the updated classifier, which can help us to decide whether we should update our classifier or not and study the influence of new data. Unless the change is unacceptably large, we can still use the original classifier. Let $\be_j \in \mR^d$ be a vector of all 0 except 1 in the $j^{th}$ element, hence we can bound the $j^{th}$ coefficient $w_{1,j}^\star$ of coefficient vector for the updated classifier using equation \eqref{eq:score} as
\begin{equation}
L(w_{1,j}^\star) \leq w_{1,j}^\star = \be_j^\top \bw_1^\star \leq U(w_{1,j}^\star).
\label{eq:sensiCoeff}
\end{equation}
In order to evaluate the tightness of the bound, we define the variable $T$ as following
\begin{equation}
T = |U(w_{1,j}^\star) - L(w_{1,j}^\star)|.
\end{equation}

\subsubsection{Task 2. Sensitivity of test instance labels}
\label{sec:sensiLabel}
Using the same framework, we can also determine the label of any test instance $\bx_i\in \mR^d$ with the updated classifier, even though we don't know the exact value for the coefficient vector of the updated linear classifier. Let $\hat y_i$ be the predicted classification result of $\bx_i$, by setting $\boldeta \coloneqq \bx_i$ in \eqref{eq:score}, we can calculate the lower and upper bounds of $\bx_i^\top \bw_1^\star$ as
\begin{equation}
L(\bx_i^\top \bw_1^\star) \leq \bx_i^\top \bw_1^\star \leq U(\bx_i^\top \bw_1^\star),
\label{eq:sensiLabel}
\end{equation}
and by using the simple facts
\begin{align}
L(\bx_i^\top \bw_1^\star) \geq 0 \Rightarrow \hat y_i = +1,  \label{eq:lowerTest}    \\
U(\bx_i^\top \bw_1^\star) \leq 0 \Rightarrow \hat y_i = -1,  \label{eq:upperTest}
\end{align}
we can obtain the classification result of $\bx_i$ without actually solving \eqref{eq:new_svmPrimal}. If the ratio of modified data set is small, we can expect that $\bw_1^\star$ won't have huge difference compared with $\bw_0^\star$, which means bounds in \eqref{eq:sensiLabel} can be sufficiently tight. 
\par In this task, we calculate the error ratio of the test instances whose lower bound and upper bound have different signs, as for these test instance, we cannot determine the classification result of $\bx_i$. The definition of this error ratio can be written as:
\begin{align}
R_{same} = n_{diff}/n_{test} \label{eq:errorratio}
\end{align}
where $n_{diff}$ is the number of instances with different signs for their estimated lower bound and upper bound; $n_{test}$ is the size of the whole test data set. By comparing error ratios for different sensitivity analysis algorithms, we can evaluate their performance.

\graphicspath{{pic/}}

\section{SENSITIVITY ANALYSIS VIA SEGMENT TEST}
In this section, we demonstrate our method for sensitivity analysis tasks. The main idea of our method is trying to restrict $\bw_1^\star$ into a region, then we can calculate the lower and upper bounds of the linear score $\boldeta^\top \bw_1^\star$ within that region. We call this kind of approach as region test as in \cite{xiang2014screening}. 

\subsection{Region Test}
\subsubsection{Sphere}
In this part, we will present the method proposed in \cite{okumura2015kdd} which serves as the lemma for this paper. 

\vspace{1ex}
\noindent
\textbf{Lemma 1 (Sphere Region)} Let $\bw_0^\star$ and $\bw_1^\star$ be the optimal solution of the problem \eqref{eq:old_svmPrimal} and \eqref{eq:new_svmPrimal} respectively. Given $\bw_0^\star$, then $\bw_1^\star$ is within a sphere region
\begin{equation}
S(\bq,r) \triangleq \{\bw_1^\star: \|\bw_1^\star - \bq\| \leq r\},
\label{eq:sphere}
\end{equation}
where $\bq$ is the center of the sphere and $r$ is the radius of the sphere. They are defined as
\begin{equation}
\bq = \frac{n_0\!+\!n_1}{2n_1}\bw_0^\star - \frac{n_\mA \!+\! n_\mS}{2Cn_1} \Delta \bs,    \label{eq:center}   
\end{equation}
\begin{equation}
r = \left\| \frac{n_\mA \!-\! n_\mS}{2n_1}\bw_0^\star + \frac{n_\mA \!+\! n_\mS}{2Cn_1} \Delta \bs \right\|,  \label{eq:radius} 
\end{equation}
where $\Delta \bs$ is
\begin{equation}
\Delta \bs = \frac{1}{n_\mA \!+\! n_\mS}\left(\sum_{i\in\mA}\nabla\ell_i(\bw_0^\star) - \sum_{i\in\mS}\nabla\ell_i(\bw_0^\star)\right).    \label{eq:deltaS}
\end{equation}
By restricting $\bw_1^\star$ into a sphere $S(\bq,r)$, we can calculate the lower and the upper bounds of $\boldeta^\top{\bw_1^\star}$ Here we introduce the following proposition proposed in \cite{okumura2015kdd}.

\vspace{1ex}
\noindent
\textbf{Proposition 1 (Sphere Test)} The lower and the upper bounds of $\boldeta^\top{\bw_1^\star}$ in the sphere region $S(\bq,r) = \{\bw_1^\star: \|\bw_1^\star - \bq\| \leq r\}\  $ are respectively

\begin{equation}
\begin{split}
L_{ST}(\boldeta^\top{\bw_1^\star}) & \triangleq \min_{{\bw_1^\star}\in S(\bq,r)} {\boldeta^\top{\bw_1^\star}} \\
&=\frac{n_0\!+\!n_1}{2n_1} \boldeta^\top\bw_0^\star - \frac{n_\mA\! +\! n_\mS}{2Cn_1} \boldeta^\top \Delta \bs \\
&-\|\boldeta\| \left\| \frac{n_\mA\!-\!n_\mS}{2n_1}\bw_0^\star + \frac{n_\mA\!+\!n_\mS}{2Cn_1} \Delta \bs \right\| ,
\end{split}
\label{eq:lb_ST}
\end{equation}
and
\begin{equation}
\begin{split}
U_{ST}(\boldeta^\top{\bw_1^\star}) \triangleq &\max_{{\bw_1^\star}\in S(\bq,r)} {\boldeta^\top{\bw_1^\star}} \\
= &\frac{n_0\!+\!n_1}{2n_1} \boldeta^\top\bw_0^\star - \frac{n_\mA \!+\! n_\mS}{2Cn_1} \boldeta^\top \Delta \bs \\
&+\|\boldeta\| \left\| \frac{n_\mA \!-\! n_\mS}{2n_1}\bw_0^\star + \frac{n_\mA \!+\! n_\mS}{2Cn_1} \Delta \bs \right\|.
\end{split}
\label{eq:ub_ST}
\end{equation}
From \eqref{eq:lb_ST} and \eqref{eq:ub_ST}, we notice that the main computation cost of the sphere test only depends on the computation of $\Delta \bs$ in \eqref{eq:deltaS}, which only involves the removed data set $n_\mS$ and the added data set $n_\mA$. Therefore, the sphere test can efficiently decrease the computing complexity. 

\subsubsection{Half Space}
Considering that $\bw_1^\star$ can be also bounded in other ways, in this part, we introduce half space test.

\vspace{1ex}
\noindent
\textbf{Theorem 1 (Half Space Region)} Given $\bw_0^\star$, then the optimal solution $\bw_1^\star$ is within in the half space
\begin{equation}
\bn^\top \bw_1^\star \leq c,
\label{eq:half1b}
\end{equation}
where $\bn$ is a unit normal vector of the plane and $c$ is the distance of the plane from the origin as in \cite{lorensen1987marching}. They are defined as
\begin{equation}
\bn = \frac{\nabla f(\bw_{C})}{||\nabla f(\bw_{C})||},
\end{equation}
\begin{equation}
c = \frac{\nabla f(\bw_{C})}{||\nabla f(\bw_{C})||} \bw_{C},
\end{equation}
where 
\begin{equation}
\bw_C \in \mR^d.
\end{equation}

\begin{proof}
Because the loss function 
\begin{equation}
f(\bw) = \frac{C}{2}\|\mathbf w\|_{2}^2+ \frac{1}{n_1}\sum_{i\in\mD_1} \ell(y_i,f(\bx_i;\bw))
\end{equation}
is convex with respect to $\bw$. According to \cite{maurer1979first}, we have
\begin{equation}
f(\bw_1^\star) \geq f(\bw_{C}) + \nabla f(\bw_{C})^\top (\bw_1^\star - \bw_{C}).
\label{eq:conv1b}
\end{equation}
As $\bw_1^\star$ is the optimal solution of the loss function, we have
\begin{equation}
f(\bw_{C}) \geq f(\bw_1^\star).
\label{eq:feasible1b}
\end{equation}
By adding \eqref{eq:conv1b} and \eqref{eq:feasible1b}, we found
\begin{equation}
\nabla f(\bw_{C})^\top (\bw_1^\star - \bw_{C}) \leq 0.
\label{eq:sumConv1b}
\end{equation}
Therefore,
\begin{equation}
\frac{\nabla f(\bw_{C})^\top}{||\nabla f(\bw_{C})||} \bw_1^\star \leq \frac{\nabla f(\bw_{C})^\top}{||\nabla f(\bw_{C})||}\bw_{C}.
\label{eq:sumConv1c}
\end{equation}
\end{proof}	

\subsubsection{Segment}
By combining Sphere Region and Half Space Region, we get following theorem.

\vspace{1ex}
\noindent
\textbf{Theorem 2 (Segment Region)} Given $\bw_0^\star$, we can find that the optimal solution $\bw_1^\star$ is included in a specific half space region $\bn^\top \bw_1^\star - c \leq 0,$
where
\begin{align}
\bn =&  \frac{(- \frac{n_\mA + n_\mS}{n_1} \Delta \bs+\frac{1}{n_1}\Delta L)}
{||- \frac{n_\mA + n_\mS}{n_1} \Delta \bs+\frac{1}{n_1}\Delta L||}; 
\label{eq:n} \\
c =& \bn^\top\!\Big(\frac{n_0}{n_1}\bw_0^\star \!-\! \frac{n_\mA \!+\! n_\mS}{Cn_1} \Delta \bs\Big);    \label{eq:c}
\end{align}
and 
\begin{equation}
\Delta L =\left(\sum_{i\in\mA}\nabla\ell_i(\bw_0^\star)+ \sum_{i\in\mS}\nabla\ell_i(\bw_0^\star)\right).
\end{equation}
Meanwhile the optimal solution $\bw_1^\star$ is also within a sphere region $\|\bw_1^\star - \bq\| \leq r.$ The sphere region will be divided into two parts by this special half space region, which means that we can restrict $\bw_1^\star$ within a smaller region called the segment region.

\begin{proof}
According to Preposition 2 and the fact that $\bw_C$ is a feasible solution of the unconstrained problem \eqref{eq:new_svmPrimal}, letting $\bw_C\!=\!\frac{n_0}{n_1}\bw_0^\star \!-\! \frac{n_\mA + n_\mS}{Cn_1} \Delta \bs$,  we have
\begin{equation}
\bn =  \frac{C\bw_C + \frac{1}{n_1}\sum\limits_{i\in\mD_1}\nabla\ell_i(\bw_C)}
{||C\bw_C + \frac{1}{n_1}\sum\limits_{i\in\mD_1}\nabla\ell_i(\bw_C)||},
\end{equation}
and
\begin{equation}
c =  \bn^\top \bw_C^\star.
\end{equation}
According the definition of $\bw_C$, we have
\begin{align}
&\frac{1}{n_1}\sum\limits_{i\in\mD_1}\nabla\ell_i(\bw_C)\notag\\
&= \frac{1}{n_1}
\sum_{i\in\mD_0} \nabla\ell_i\big(\frac{n_0}{n_1}\bw_0^\star-\frac{n_\mA + n_\mS}{Cn_1} \Delta \bs\big) \notag\\
&+ \frac{1}{n_1}
\sum_{i\in\mD_A} \nabla\ell_i\big(\frac{n_0}{n_1}\bw_0^\star-\frac{n_\mA + n_\mS}{Cn_1} \Delta \bs\big)\notag\\
&+\frac{1}{n_1}
\sum_{i\in\mD_S} \nabla\ell_i\big(\frac{n_0}{n_1}\bw_0^\star-\frac{n_\mA + n_\mS}{Cn_1} \Delta \bs\big).
\label{eq:proof2.1}	
\end{align}
By using quadratic approximation of Taylor's Formula as in \cite{hummel1949generalization}, we have
\begin{align}
&\nabla\ell_i\big(\frac{n_0}{n_1}\bw_0^\star-\frac{n_\mA + n_\mS}{Cn_1} \Delta \bs\big) \notag\\
&= \nabla\ell_i\big(\bw_0^\star - \frac{n_A + n_B}{n_1}\bw_0^\star-\frac{n_\mA + n_\mS}{Cn_1} \Delta \bs\big) \notag\\
&\approx \nabla\ell_i\big(\bw_0^\star) + \nabla^2\ell_i\big(\bw_0^\star)(-\frac{n_A + n_B}{n_1}\bw_0^\star-\frac{n_\mA + n_\mS}{Cn_1} \Delta \bs)\notag\\
&+ \nabla^3\ell_i\big(\bw_0^\star)(-\frac{n_A + n_B}{n_1}\bw_0^\star-\frac{n_\mA + n_\mS}{Cn_1} \Delta \bs)^2\notag\\
&\approx \nabla\ell_i\big(\bw_0^\star).
\label{eq:proof2.2}	
\end{align}

Moreover, $\bw_0^\star$ is the optimal solution of the convex loss function  $f(\bw) = \frac{C}{2}\|\mathbf w\|_{2}^2+ \frac{1}{n_0}\sum_{i\in\mD_0} \ell(y_i,f(\bx_i;\bw))$. Thus, $\bw_0^\star$ meets following equation:
\begin{equation}
C \mathbf w^\top+ \frac{1}{n_0}\sum_{i\in\mD_0} \nabla \ell(y_i,f(\bx_i;\bw)) = 0,
\end{equation}
which is equal to
\begin{equation}
\sum_{i\in\mD_0} \nabla \ell(y_i,f(\bx_i;\bw)) = - n_0 C \mathbf w^\top.
\label{eq:proof2.3}	
\end{equation}
By combining the equation \eqref{eq:proof2.1}, \eqref{eq:proof2.2} and \eqref{eq:proof2.3}, we have
\begin{align}
&\frac{1}{n_1}
\sum_{i\in\mD_1} \nabla\ell_i\big(\bw_C\big) \notag\\
&= \frac{1}{n_1}
\sum_{i\in\mD_0} \nabla\ell_i\big(\bw_0^\star) +
\frac{1}{n_1}
\sum_{i\in\mD_A} \nabla\ell_i\big(\bw_0^\star)+	
\frac{1}{n_1}
\sum_{i\in\mD_S} \nabla\ell_i\big(\bw_0^\star)	\notag\\
 &= - C\frac{n_0}{n_1}\bw_0^\star+
\Delta L.
\end{align}

\begin{figure}
	\begin{center}
		\includegraphics[width=0.3\textwidth]{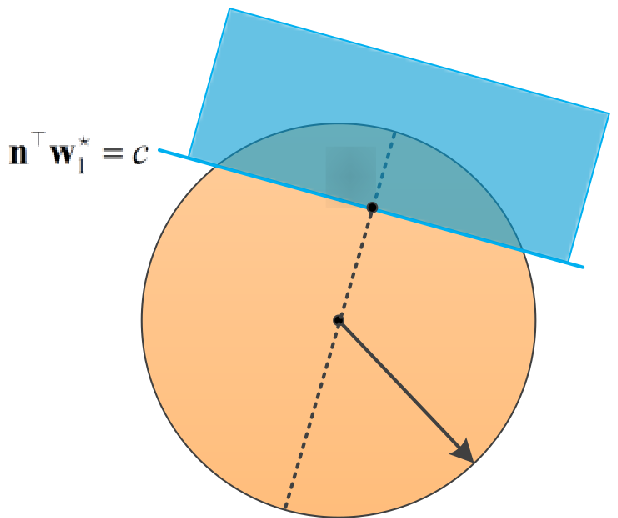}
	\end{center}
	\caption{Illustration of Segment Region $D(\bq,r;\bn,c)$.}
	\label{fig:segmentRegion}
\end{figure}

After acquired the half space region, we consider a segment region test based on nonempty intersection of the closed sphere $\{\bw_1^\star: \|\bw_1^\star - \bq\| \leq r\}$ and the closed half space $\{\bw_1^\star: \bn^\top \bw_1^\star \leq c\}$. As in \cite{boyer1994influence}, the distance between the sphere center $\bq$ and the plane can written as
\begin{equation}
\psi r =\bn^\top \bq - c.
\end{equation}
Thus, the coefficient $\psi$ and the intersection point $\bq_d$ between the plane and the radius of the sphere which is orthogonal to the plane can be defined as 

\begin{equation}
\begin{split}
\psi =& \frac{\bn^\top \bq - c}{r},     \\
\bq_d =& \bq - \psi r \bn,    \\
\end{split}
\label{eq:notation}
\end{equation}
where $\bq$, $r$, $\bn$ and $c$ are defined in \eqref{eq:center}, \eqref{eq:radius}, \eqref{eq:n} and \eqref{eq:c}. We have
\begin{equation}
\psi =\frac{ \bn^\top \Big( \frac{n_\mA - n_\mS}{2n_1}\bw_0^\star + \frac{n_\mA + n_\mS}{2Cn_1} \Delta \bs \Big) }
{\Big\| \frac{n_\mA - n_\mS}{2n_1}\bw_0^\star + \frac{n_\mA + n_\mS}{2Cn_1} \Delta \bs \Big\| }.
\label{eq:psi}
\end{equation}
So, the range of $\psi$ is $[-1,1]$, which means that the intersection point $\bq_d$ is located inside the sphere. Thus, the sphere region is divided by the half space region into two parts and the size of the generated segment region is smaller than the sphere region. In this way, we restrict $\bw_1^\star$ into a smaller segment region. Figure \ref{fig:segmentRegion} illustrates the segment region $D(\bq,r;\bn,c) = \{ \bw_1^\star: \|\bw_1^\star - \bq\| \leq r,\ \bn^\top \bw_1^\star \leq c\}$. 
\end{proof}

\vspace{1ex}
\noindent
\textbf{Theorem 3 (Segment Test)} Using the definition of $\bq$, $r$, $\bn$ and $c$ in \eqref{eq:center}, \eqref{eq:radius}, \eqref{eq:n} and \eqref{eq:c}, we can found the upper and the lower bound of $\boldeta^\top \bw_1^\star$ and the computing complexity of this calculation only depends on the added and removed data set $n_\mA$ and $n_\mS$. The lower and the upper bound of $\boldeta^\top \bw_1^\star$ in the segment region $D(\bq,r;\bn,c) = \{ \bw_1^\star: \|\bw_1^\star - \bq\| \leq r,\ \bn^\top \bw_1^\star \leq c\}$ are respectively
\begin{equation}
\begin{split}
L_{DT} \!\triangleq\! & \min_{{\bw_1^\star}\in D}  \boldeta^\top \bw_1^\star \\
\!=\!&
\begin{cases}
\bq^\top \boldeta \!-\! r\|\boldeta\|  & t \!>\! \psi \|\boldeta \| ,   \\
\bq^\top \boldeta \!-\! \psi r t \!-\! r \sqrt{1 \!-\! \psi^2} \sqrt{\|\boldeta\|^2 \!-\! t^2} & t \!\leq\! \psi \|\boldeta \| ,\\
\end{cases}
\end{split}
\label{eq:lb_DT}
\end{equation}
and
\begin{equation}
\begin{split}
U_{DT} \triangleq & \max_{{\bw_1^\star}\in D} \boldeta^\top \bw_1^\star \\
=&
\begin{cases}
\bq^\top \boldeta \!+\! r\|\boldeta\|  \ & t \!<\! -\psi \|\boldeta \|,    \\
\bq^\top \boldeta \!-\! \psi r t \!+\! r \sqrt{1 \!-\! \psi^2} \sqrt{\|\boldeta\|^2 \!-\! t^2} \ & t \!\geq\! -\psi \|\boldeta \|, \\
\end{cases}
\end{split}
\label{eq:ub_DT}
\end{equation}
where
$$ t \triangleq \bn^\top \boldeta. $$

\begin{proof}
	We can obtain the lower bound of $\boldeta^\top \bw_1^\star$ by solving the following optimization problem
	\begin{equation}
	\begin{split}
	\min_{ \bw_1^\star}\quad& \boldeta^\top \bw_1^\star \\
	\textrm{s.t.}\quad& (\bw_1^\star - \bq)^\top (\bw_1^\star - \bq) \leq r^2 \\
	& \bn^\top \bw_1^\star - c \leq 0.
	\end{split}
	\label{eq:opt1}
	\end{equation}
	The Lagrange function \cite{borneas1959generalization} of \eqref{eq:opt1} can be written as
	\begin{equation}
	\mL( \bw_1^\star , \mu, \sigma) \!=\! \boldeta^\top \bw_1^\star \!+\! \mu \left((\bw_1^\star \!-\! \bq)^\top (\bw_1^\star \!-\! \bq) \!-\! r^2\right) \!+\! \sigma(\bn^\top \bw_1^\star \!-\! c),
	\label{eq:lag1}
	\end{equation}
	where $\mu \geq 0$ and $\sigma \geq 0$ are Lagrange multipliers. As in \cite{wu2007karush}, setting the derivative with regard to primal variable $\bw_1^\star$ to zero yields
	\begin{equation}
	\frac{\partial \mL}{\partial \bw_1^\star} \!=\! \boldeta + 2\mu(\bw_1^\star-\bq) + \sigma \bn=\bzero    \Rightarrow \bw_1^\star = \bq - \frac{\boldeta + \sigma\bn}{2\mu}.
	\label{eq:optW}
	\end{equation}
	Substituting $\bw_1^\star$ into $\boldeta^\top \bw_1^\star$,
	\begin{equation}
	\min_{ \bw_1^\star}\ \boldeta^\top \bw_1^\star = \bq^\top \boldeta - \frac{\| \boldeta \|^2}{2\mu} - \frac{\sigma \bn^\top \boldeta }{2\mu}.
	\label{eq:min}
	\end{equation}
	Substituting $\bw_1^\star$ into \eqref{eq:lag1},
	\begin{equation}
	\mL( \mu, \sigma) 
	= \bq^\top\boldeta - \frac{\|\boldeta \|^2}{4\mu} - \frac{\sigma t }{2\mu} - \frac{\sigma^2}{4\mu} - \mu r^2 + \sigma \psi r,
	\label{eq:lag2}
	\end{equation}
	where $\psi$ is defined in \eqref{eq:psi}, $ t = \bn^\top \boldeta $ and $\| \bn \| = 1$.
	Hence, we can divide this problem into two cases: 
	
	(a) $\bn^\top \bw_1^\star < c$
	\par In order to satisfy the complementary slackness condition $\sigma(\bn^\top \bw_1^\star \!-\! c)=0$ in \eqref{eq:lag1}, we found
	\begin{align}
	\sigma = 0.
	\end{align}
	Then substitute $\sigma = 0$ into \eqref{eq:lag2} and set the derivative of $\mL( \mu, \sigma)$ of \eqref{eq:lag2} with respect to $\mu$ equal to zero, we have
	\begin{align}
	\mu = \frac{\|\boldeta \|}{2r}.
	\end{align}
	Based on \eqref{eq:optW}, it's simple to check out that
	\begin{equation}
	\begin{split}
	\bn^\top \bw_1^\star - c 
	=& \frac{r}{\|\boldeta \|} (\psi \|\boldeta \| - t).
	\label{eq:cond1}
	\end{split}
	\end{equation}
	Therefore $\bn^\top \bw_1^\star < c$ is equal to $t > \psi \|\boldeta \|$. Substitute $\sigma$ and $\mu$ into \eqref{eq:min}, we obtain
	\begin{equation}
	\min_{ \bw_1^\star}\ \boldeta^\top \bw_1^\star = \bq^\top \boldeta - r\|\boldeta\|.
	\label{eq:minCase1}
	\end{equation}
	
	(b) $\bn^\top \bw_1^\star = c$	
	\par Setting the derivatives of
	$\mL( \mu, \sigma)$ of \eqref{eq:lag2} with respect to $\mu$ and $\sigma$ equal to zero yields, we found
	\begin{align}
	& \|\boldeta\|^2 + 2\sigma t + \sigma^2 = 4\mu^2 r^2, \label{eq:sigma} \\ 
	& \sigma = 2\mu r \psi - t. \label{eq:mu}
	\end{align}
	Using \eqref{eq:sigma} and \eqref{eq:mu} together, we can easily get
	\begin{align}
	&\sigma = -t + \psi \sqrt{\frac{\|\boldeta\|^2 - t^2}{1-\psi^2} }, \\
	&\mu = \frac{1}{2r} \sqrt{\frac{\|\boldeta\|^2 - t^2}{1-\psi^2} }.
	\end{align}
	According to the Karush–Kuhn–Tucker conditions, we have
	\begin{equation}
	\sigma \geq 0.
	\end{equation}
    Thus,
	\begin{equation}
	\psi \sqrt{\frac{\|\boldeta\|^2 - t^2}{1-\psi^2} } - t \geq 0.   \label{eq:cond2}
	\end{equation}
	It's simple to prove that \eqref{eq:cond2} is equal to $t \leq \psi \|\boldeta \|$. Substitute $\sigma$ and $\mu$ into \eqref{eq:min}
	\begin{equation}
	\begin{split}
	\min_{ \bw_1^\star}\ \boldeta^\top \bw_1^\star
	=&\bq^\top \boldeta \!-\! \psi rt \!-\! r \sqrt{1 \!-\! \psi^2} \sqrt{\|\boldeta\|^2 \!-\! t^2}.
	\label{eq:minCase2}
	\end{split}
	\end{equation}
	By combining case (a) and case (b), the lower bound in \eqref{eq:lb_DT} is obtained. And the upper bound \eqref{eq:ub_DT} can be similarly derived. 
\end{proof}

\subsection{Sensitivity analysis tasks}
\subsubsection{Sensitivity of coefficients}
By using segment test, we can get the lower and upper bound of the $j^{th}$ element $w_{1,j}^\star$ of the coefficient vector $\bw_1^\star$ for the updated linear classifier by substituting $\boldeta \coloneqq \be_j$ in \eqref{eq:lb_DT} and \eqref{eq:ub_DT}.

\vspace{1ex}
\noindent
\textbf{Corollary 1 (Sensitivity of coefficients)} $\forall j \in [1,d] $ , the $j^{th}$ coefficient $w_{1,j}^\star$ for coefficient vector of the updated classifier satisfies
\begin{equation}
\begin{split}
& L_{DT}(w_{1,j}^\star) \\
\!=\!&
\begin{cases}
\frac{n_0\!+\!n_1}{2n_1} w_{0,j}^\star \!-\! \frac{n_\mA\! +\! n_\mS}{2Cn_1} \Delta s_j \!-\! r & t \!>\! \psi,    \\
\frac{n_0\!+\!n_1}{2n_1} w_{0,j}^\star \!-\! \frac{n_\mA\! +\! n_\mS}{2Cn_1} \Delta s_j \!-\! \psi r  t \!-\! r \sqrt{1 \!-\! \psi^2} \sqrt{1 \!-\! t^2} & t \!\leq\! \psi,
\end{cases}
\end{split}
\label{eq:lbCoeff}
\end{equation}
and
\begin{equation}
\begin{split}
& U_{DT}(w_{1,j}^\star) \\
\!=\!&
\begin{cases}
\frac{n_0\!+\!n_1}{2n_1} w_{0,j}^\star \!-\! \frac{n_\mA\! +\! n_\mS}{2Cn_1} \Delta s_j \!+\! r & t \!<\! -\psi,    \\
\frac{n_0\!+\!n_1}{2n_1} w_{0,j}^\star \!-\! \frac{n_\mA\! +\! n_\mS}{2Cn_1} \Delta s_j \!-\! \psi r  t \!+\! r \sqrt{1 \!-\! \psi^2} \sqrt{1 \!-\! t^2} & t \!\geq\! -\psi,
\end{cases}
\end{split}
\label{eq:ubCoeff}
\end{equation}
where $\Delta s_j$ are the $j^{th}$ coefficient of $\Delta \bs$ defined in \eqref{eq:deltaS}; $r$ and $\psi$ are defined in \eqref{eq:radius} and \eqref{eq:psi} respectively; $ t = \bn^\top \be_j$.
\par Given lower and upper bounds of $w_{1,j}^\star$, we can obtain the bounds for $\bw_1^\star$. Here we choose the average for the tightness of the bound of each instance
$\frac{1}{d}\sum_{j=1}^d(T_j)$ as the evaluation variable called bound tightness.

\subsubsection{Sensitivity of test instance labels}
Next, we use \textbf{Theorem 3} for the sensitivity of test instance labels. Substituting $\boldeta \coloneqq \bx_i$ into \eqref{eq:lb_DT} and \eqref{eq:ub_DT}, we can get the lower and upper bounds of $\bx_i^\top \bw_1^\star$. If the signs of the lower and upper bound are the same, we can infer the test instance label by calculating the bounds instead of calculating the coefficient vector $\bw_1^\star$ of the updated linear classifier.

\vspace{1ex}
\noindent
\textbf{Corollary 2 (Sensitivity of test instance labels)} For any test instance $\bx_i$, the classification result using the updated classifier $\bw_1^\star$ is
\begin{equation}
\hat y_i \coloneqq \text{sgn}(f(\bx_i^\top \bw_1^\star)),
\label{eq:testLabel1}
\end{equation}
which satisfies
\begin{equation}
\hat y_i =
\begin{cases}
+1  &  L_{DT}(\bx_i^\top \bw_1^\star) \geq 0,  \\
-1  &  U_{DT}(\bx_i^\top \bw_1^\star) \leq 0,  \\
unknown & otherwise,
\end{cases}
\label{eq:testLabel2}
\end{equation}
where
\begin{equation}
\begin{split}
&L_{DT}(\bx_i^\top \bw_1^\star) \\
\!=\!&
\begin{cases}
\bq^\top \bx_i \!-\! r\|\bx_i\|  & t \!>\! \psi \|\bx_i \| ,   \\
\bq^\top \bx_i \!-\! \psi r t \!-\! r \sqrt{1 \!-\! \psi^2} \sqrt{\|\bx_i\|^2 \!-\! t^2} & t \!\leq\! \psi \|\bx_i \| ,\\
\end{cases}
\end{split}
\label{eq:lbLabel}
\end{equation}
and
\begin{equation}
\begin{split}
&U_{DT}(\bx_i^\top \bw_1^\star) \\
=&
\begin{cases}
\bq^\top \bx_i \!+\! r\|\bx_i\|  \ & t \!<\! -\psi \|\bx_i \|,    \\
\bq^\top \bx_i \!-\! \psi r t \!+\! r \sqrt{1 \!-\! \psi^2} \sqrt{\|\bx_i\|^2 \!-\! t^2} \ & t \!\geq\! -\psi \|\bx_i \|, \\
\end{cases}
\end{split}
\label{eq:ubLabel}
\end{equation}
$\bq$, $r$ and $\psi$ are defined in \eqref{eq:center}, \eqref{eq:radius} and \eqref{eq:psi} respectively, $ t = \bn^\top \bx_i$.

\section{Experimental Results}
In this section, we conduct extensive experiments to evaluate our method on real-world data sets. We first provide a brief description of data sets and experimental setup, then we show the simulation result of four simulations and evaluate the performance of different algorithms.

\begin{table}
	\centering
	\caption{Data sets used for experiments}\label{DataBound}
	\begin{tabular}{llllll}
		\hline
		&Data set & $n_{train}$ & $dim$ & $n_{test}$ & $experiment$\\
		\noalign{\smallskip}\hline\noalign{\smallskip}	
		D1 & w8a     & 49749    & 300   & 14951   & I \\
		D2 & a9a     & 32561    & 123   & 16281   & II \\
		D3 & cod-rna & 59535    & 8     & 271617  & III \& IV \\
		\hline
	\end{tabular}
\end{table}

\paragraph{Data sets}
In this part, we use three data sets from LIBSVM data set repository \cite{chang2011libsvm} for comparison, and data sets used for each test are presented in Table~\ref{DataBound}. For each test, the regularization constant $C$ is set to be \{0.2, 0.5, 1\}, meanwhile the ratio of modified data set $P_{up}$ is set to be \{0.01\%, 0.02\%, 0.05\%,0.1\%, 0.2\%, 0.5\%, 1\%, 2\%, 5\%, 10\%\}. 

\paragraph{experiment setting}
We use the Matlab linear system proposed in \cite{chen1998linear} to calculate relevant bounds or values for the coefficient vector of the linear classifier with the L2-SVM or L2-logistic regression. In experiments, we compare our method with the algorithm described in \cite{okumura2015kdd}. Our method and the algorithm proposed by \cite{okumura2015kdd} are called segment test and sphere test respectively. In order to guarantee the reliability of the experiment, for each set with a different ratio of modified data set $P_{up}$, different regularization constant $C$ and different data set, we repeat the experiment for 30 times. All the experiments were performed on a desktop with 4-core Intel(R) Core(TM) processors of 2.80 GHz and 8 GB of RAM. The Matlab 2017 is used to conduct the simulation and the R is used for the data visualization.
 
\subsection{Results on sensitivity of coefficients task}
\begin{figure*}
\setlength{\tabcolsep}{10pt} 
\setlength{\abovecaptionskip}{0.cm}
\setlength{\belowcaptionskip}{-0.3cm}
	\begin{center}
		\includegraphics[width=1\textwidth]{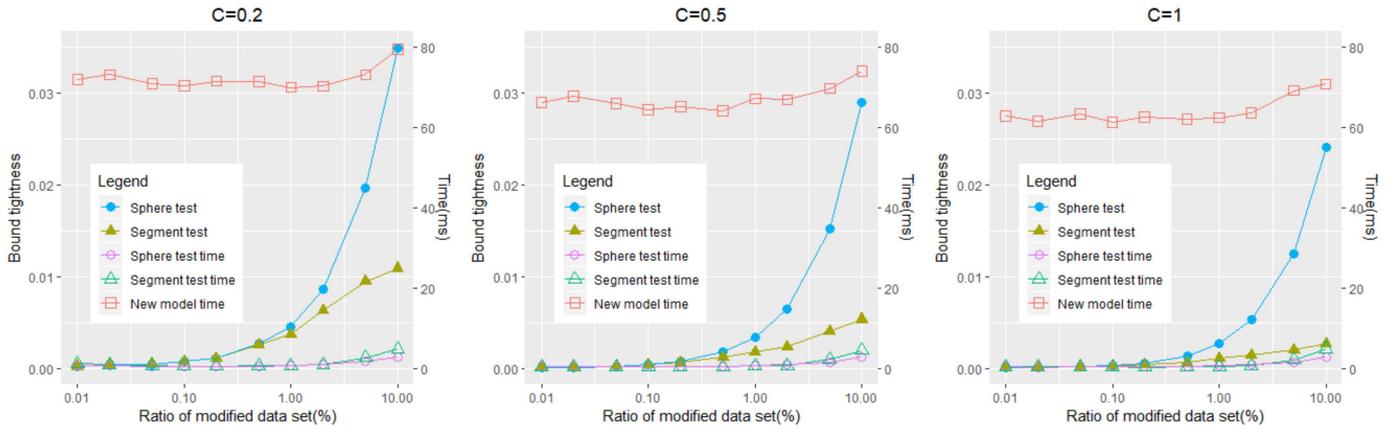}
	\end{center}
	\caption{Comparison of coefficients bounds with L2-SVM. Lower y values mean that smaller differences between the upper and lower bounds for coefficients.}
	\label{fig:svmbound}
\end{figure*}

\begin{table*}[htbp]
\begin{center}
	\caption{Results for sensitivity of coefficients experiment with L2-SVM}\label{Sensitivity_SVM}
\scalebox{0.85}{\begin{tabular}{|c|c||c|c|c||c|c|c||c|c|c||c|}
	\hline
	\multicolumn{2}{|c||}{\multirow{2}{*}{}} &
	\multicolumn{10}{|c|}{$C = 0.1$} \\
	\cline{1-12}
	\multicolumn{2}{|c||}{$P_{up}(\%)$}&0.01&0.02&0.05&0.1&0.2&0.5&1&2&5&10\\
	\hline
	New Model&Time(ms)&72.03&73.26&70.96&70.49&71.60&71.56&70.06&70.38&73.24&79.60 \\
	\hline
	\multirow{2}{*}{Sphere Test}&Time(ms)&0.67&0.58&0.46&0.43&0.41&0.52&0.62&0.87&1.62&2.78 \\
	\cline{2-12}
	&Tightness&2.03e-4&3.74e-4&4.54e-4&7.40e-4&1.10e-3&2.69e-3&4.50e-3&8.66e-3&1.97e-2&3.50e-2 \\
	\hline
	\multirow{2}{*}{Segment Test}&Time(ms)&1.23&0.95&0.65&0.51&0.50&0.58&0.55&0.86&2.58&4.78 \\
	\cline{2-12}
	&Tightness&2.03e-4&3.74e-4&4.54e-4&7.40e-4&1.10e-3&2.55e-3&3.73e-3&6.29e-3&9.46e-3&1.09e-2 \\
	\hline
	\multicolumn{2}{|c||}{\multirow{2}{*}{}} &
	\multicolumn{10}{|c|}{$C = 0.5$} \\
	\hline
	New Model&Time(ms)&66.33&67.92&66.13&64.50&65.23&64.26&67.29&67.01&69.77&74.01 \\
	\hline
	\multirow{2}{*}{Sphere Test}&Time(ms)&0.39&0.36&0.37&0.35&0.39&0.46&0.66&0.87&1.56&2.78 \\
	\cline{2-12}
	&Tightness&6.45e-5&7.20e-5&2.23e-4&3.79e-4&7.26e-4&1.73e-3&3.44e-3&6.41e-3&1.52e-2&2.90e-2 \\
	\hline
	\multirow{2}{*}{Segment Test}&Time(ms)&0.42&0.35&0.37&0.36&0.38&0.39&0.60&0.71&2.23&4.43 \\
	\cline{2-12}
	&Tightness&6.45e-5&7.15e-5&2.23e-4&3.61e-4&6.14e-4&1.19e-3&1.77e-3&2.36e-3&4.04e-3&5.33e-3 \\
	\hline
	\multicolumn{2}{|c||}{\multirow{2}{*}{}} &
	\multicolumn{10}{|c|}{$C = 1$} \\
	\hline
	New Model&Time(ms)&62.89&61.66&63.48&61.38&62.64&62.10&62.49&63.67&69.30&70.94 \\
	\hline
	\multirow{2}{*}{Sphere Test}&Time(ms)&0.31&0.28&0.29&0.31&0.33&0.41&0.55&0.81&1.48&2.71 \\
	\cline{2-12}
	&Tightness&3.25e-5&7.92e-5&1.60e-4&2.76e-4&5.46e-4&1.31e-3&2.74e-3&5.36e-2&1.25e-2&2.41e-2 \\
	\hline
	\multirow{2}{*}{Segment Test}&Time(ms)&0.36&0.31&0.30&0.30&0.29&0.36&0.42&0.66&2.03&4.88 \\
	\cline{2-12}
	&Tightness&3.25e-5&7.78e-5&1.60e-4&2.28e-4&3.91e-4&6.16e-4&1.05e-3&1.40e-3&1.95e-3&2.65e-3 \\
	\hline
\end{tabular}}
\end{center}
\end{table*}

\begin{figure*}
\setlength{\tabcolsep}{10pt} 
\setlength{\abovecaptionskip}{0.cm}
\setlength{\belowcaptionskip}{-0.3cm}
	\begin{center}
		\includegraphics[width=1\textwidth]{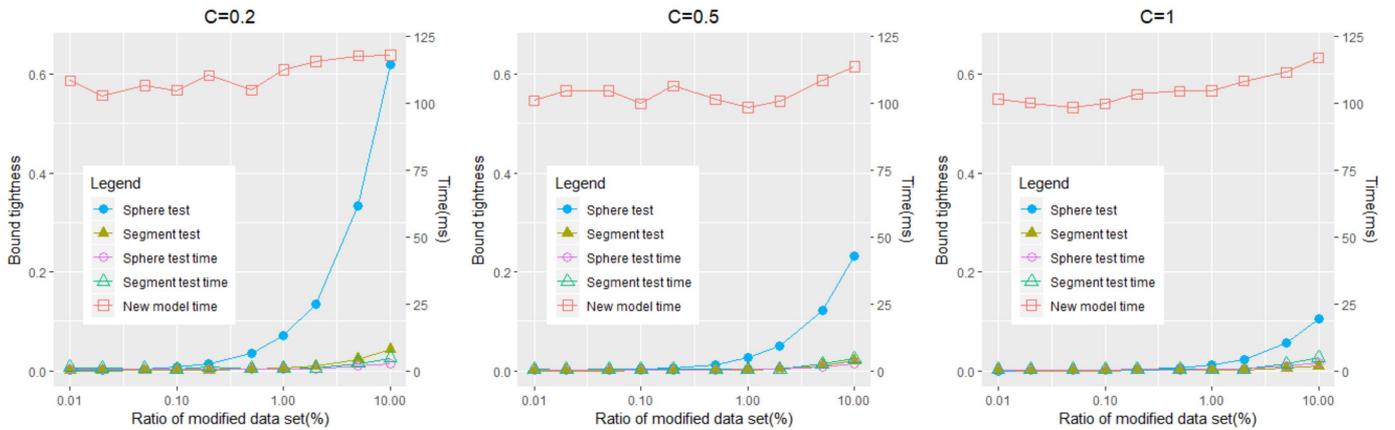}
	\end{center}
	\caption{Comparison of coefficients bounds with L2 logistic regression. Lower y values mean that smaller differences between the upper and lower bounds for coefficients are smaller.}
	\label{fig:boundlogit}
\end{figure*}

\begin{table*}[htbp]
	\begin{center}
		\caption{Results for sensitivity of coefficients experiment with L2-Logistic Regression}\label{Sensitivity_Log}
		\scalebox{0.85}{\begin{tabular}{|c|c||c|c|c||c|c|c||c|c|c||c|}
			\hline
			\multicolumn{2}{|c||}{\multirow{2}{*}{}} &
			\multicolumn{10}{|c|}{$C = 0.1$} \\
			\cline{1-12}
			\multicolumn{2}{|c||}{$P_{up}(\%)$}&0.01&0.02&0.05&0.1&0.2&0.5&1&2&5&10\\
			\hline
			New Model&Time(ms)&108.33&102.77&106.53&104.72&110.22&104.83&112.31&115.49&117.58&117.83 \\
			\hline
			\multirow{2}{*}{Sphere Test}&Time(ms)&0.80&0.74&0.54&0.53&0.79&0.49&0.58&0.82&1.73&2.60 \\
			\cline{2-12}
			&Tightness&8.60e-4&1.81e-3&4.05e-3&6.94e-3&1.49e-2&3.43e-2&7.03e-2&1.35e-1&3.34e-1&6.20e-1 \\
			\hline
			\multirow{2}{*}{Segment Test}&Time(ms)&1.14&0.88&0.73&0.63&1.42&0.72&0.77&0.86&2.76&4.67 \\
			\cline{2-12}
			&Tightness&5.63e-5&1.09e-4&2.82e-4&4.88e-4&1.01e-3&2.36e-3&4.88e-3&9.55e-3&2.30e-2&4.35e-2 \\
			\hline
			\multicolumn{2}{|c||}{\multirow{2}{*}{}} &
			\multicolumn{10}{|c|}{$C = 0.5$} \\
			\hline
			New Model&Time(ms)&100.94&104.56&104.64&99.75&106.38&101.31&98.30&100.66&108.43&113.54 \\
			\hline
			\multirow{2}{*}{Sphere Test}&Time(ms)&0.40&0.39&0.39&0.40&0.41&0.41&0.48&0.65&1.57&2.51 \\
			\cline{2-12}
			&Tightness&3.66e-4&6.58e-4&1.92e-3&2.61e-3&5.08e-3&1.27e-2&2.68e-2&4.92e-2&1.21e-1&2.31e-1 \\
			\hline
			\multirow{2}{*}{Segment Test}&Time(ms)&0.50&0.43&0.45&0.49&0.55&0.52&0.65&0.67&2.44&4.47 \\
			\cline{2-12}
			&Tightness&3.33e-5&6.24e-5&1.45e-4&2.40e-4&4.31e-4&1.11e-3&2.23e-3&4.20e-3&1.03e-2&1.98e-2 \\
			\hline
			\multicolumn{2}{|c||}{\multirow{2}{*}{}} &
			\multicolumn{10}{|c|}{$C = 1$} \\
			\hline
			New Model&Time(ms)&101.46&100.02&98.24&99.69&103.25&104.37&104.60&108.21&111.47&116.87 \\
			\hline
			\multirow{2}{*}{Sphere Test}&Time(ms)&0.39&0.43&0.43&0.40&0.44&0.43&0.61&0.75&1.70&2.79 \\
			\cline{2-12}
			&Tightness&1.95e-4&3.51e-4&8.12e-4&1.37e-3&2.52e-3&5.44e-3&1.18e-2&2.29e-2&5.57e-2&1.05e-2 \\
			\hline
			\multirow{2}{*}{Segment Test}&Time(ms)&0.33&0.36&0.38&0.36&0.39&0.48&0.61&0.67&2.56&4.82 \\
			\cline{2-12}
			&Tightness&1.96e-5&3.71e-5&7.47e-5&1.41e-4&2.57e-4&5.14e-4&1.07e-3&1.98e-3&4.77e-3&9.21e-3 \\
			\hline
		\end{tabular}}
	\end{center}
\end{table*}

\subsubsection{L2-SVM}
Here we show the results for the sensitivity of coefficients task described in Section \ref{sec:sensiCoeff}. Comparison with respect to the bound tightness and the computing time between sphere test and segment test with L2-SVM can be found in Figure \ref{fig:svmbound}. And detailed values can found in Table~\ref{Sensitivity_SVM}. In this test, we found:

When $C$ increases, the bound tightness for both tests decreases, while the difference between the bound tightness of two tests becomes larger. More specifically, when $C$ is equal to $0.2$ and $P_{up}$ is equal to $10\%$, the bound tightness of the sphere test is almost $3$ times of the bound tightness of the segment test.
When $C$ is equal to $0.5$ and $P_{up}$ is equal to $10\%$, the bound tightness of the sphere test is almost $5$ times of the bound tightness of the segment test.
When $C$ is equal to $1$ and $P_{up}$ is equal to $10\%$, the bound tightness of the sphere test is almost $10$ times of the bound tightness of the segment test.

With the same regularization constant $C$, when the ratio of modified data set $P_{up}$ is inferior to $0.1\%$, the performances of two tests almost have no difference. 
However, the advantage of the segment test becomes more and more pronounced when the $P_{up}$ continues to increase. 

In general, the computing time of the segment test is longer than the computing time of the sphere test, while the difference can be ignored. Plus, Compared with training a new classifier from scratch, the computing time of the segment test is still very short.

\subsubsection{L2-Logistic Regression}
Comparison with respect to the bound tightness and the computing time between sphere test and segment test of coefficients bounds with L2-Logistic Regression can be found in Figure \ref{fig:boundlogit}. And detailed values can found in Table~\ref{Sensitivity_Log}. We found:

When $C$ increases, the bound tightness for both tests decreases faster than their bound tightness with L2-SVM, and the difference between the bound tightness of two algorithms becomes smaller, which is the difference from the trend with L2-SVM. When $C$ is equal to $0.2$ and $P_{up}$ is equal to $10\%$, the bound tightness of the sphere test is almost $14$ times of the bound tightness of the segment test.
When $C$ is equal to $0.5$ and $P_{up}$ is equal to $10\%$, the bound tightness of the sphere test is almost $12$ times of the bound tightness of the segment test.
When $C$ is equal to $1$ and $P_{up}$ is equal to $10\%$, the bound tightness of the proposed algorithm is almost one-eleventh of the algorithm in \cite{okumura2015kdd}.
the bound tightness of the sphere test is almost $11$ times of the bound tightness of the segment test.

This experiment shares the second conclusion of the first experiment. In this experiment, the time of training a new model is longer than that of last experiment by using the different function in 'liblinear-2.21', while the computing time of segment test and that of the sphere test don't have evident difference with their computing time in the first experiment, as in these two tests, they only use the nature for the convexity of the loss function without solving relevant optimization problems. 

From the analysis above, we found, in the sensitivity analysis task of coefficients, the segment test performs better than the sphere test in the aspect of accuracy without evidently increasing computing complexity.

\subsection{Results on sensitivity of test instance labels task}
\subsubsection{L2-SVM}
\begin{figure*}
\setlength{\tabcolsep}{10pt} 
\setlength{\abovecaptionskip}{0.cm}
\setlength{\belowcaptionskip}{-0.3cm}
	\begin{center}
		\includegraphics[width=1\textwidth]{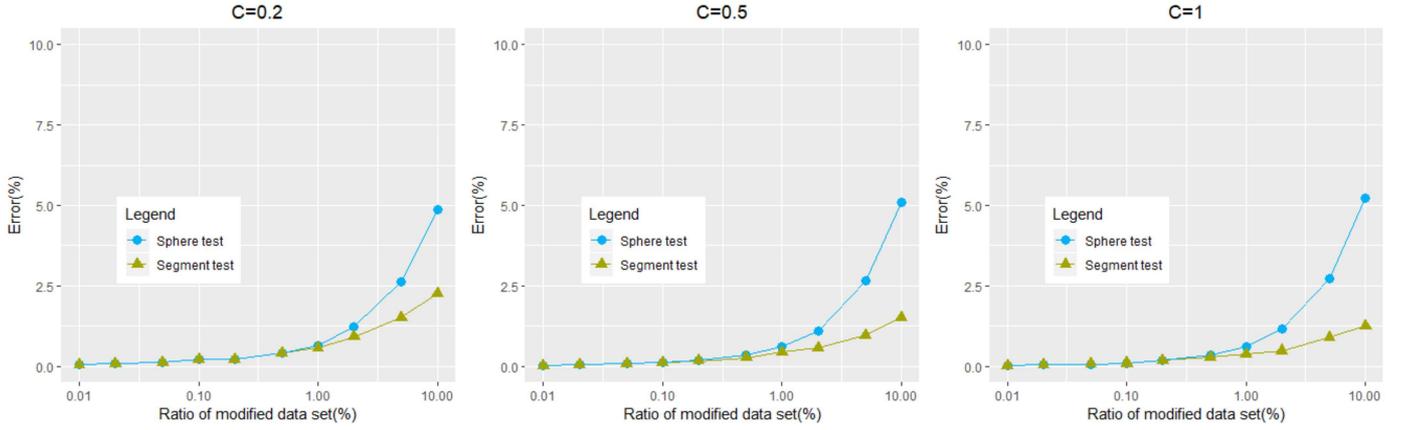}
	\end{center}
	\caption{Comparison of coefficients bounds with L2-SVM. The Error ratio close to 0 means that the bounds are tighter.}
	\label{fig:svmlabel}
\end{figure*}

Here we show results for sensitivity of test instance labels task described in Section \ref{sec:sensiLabel}. A comparison in respect of the error ratio $R_{same}$ between sphere test and segment test with L2-SVM can be found in Figure \ref{fig:svmlabel} and detailed values can be found in Table~\ref{Sensitivity2_SVM}. The reason for which we chose to use data set cod-rna in this test is that in data set cod-rna, the test set is larger than its training set. We found:

When $C$ increases, the error ratio for proposed algorithm decreases, while the error ratio of the algorithm in \cite{okumura2015kdd} doesn't have an evident trend. The difference between the error ratio of the two tests also increases. 

When $C$ is equal to $0.2$ and $P_{up}$ is equal to $10\%$, the error ratio for the sphere test is almost $2$ times of the error ratio for the segment test.
When $C$ is equal to $0.5$ and $P_{up}$ is equal to $10\%$,  the error ratio for the sphere test is almost $3$ times of the error ratio for the segment test.
When $C$ is equal to $1$ and $P_{up}$ is equal to $10\%$,  the error ratio for the sphere test is almost $4$ times of the error ratio for the segment test.

With the same regularization constant $C$, the performances of the two tests almost have no difference when the ratio of modified data set $P_{up}$ is inferior to $0.5\%$. However, the advantage of the segment test becomes more and more evident, when the $P_{up}$ continues to increase. 

\subsubsection{L2-Logistic Regression}
\begin{figure*}
\setlength{\tabcolsep}{10pt} 
\setlength{\abovecaptionskip}{0.cm}
\setlength{\belowcaptionskip}{-0.3cm}
	\begin{center}
		\includegraphics[width=1\textwidth]{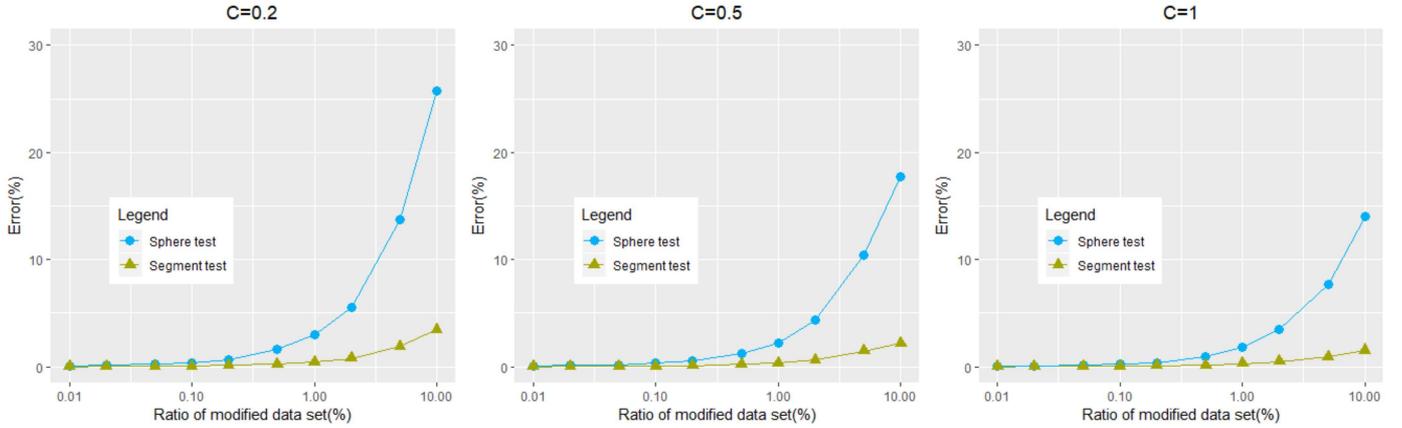}
	\end{center}
	\caption{Comparison of coefficients bounds with L2 logistic regression. The Error ratio close to 0 means that the bounds are tighter.}
	\label{fig:logistlabel}
\end{figure*}

Comparison with respect to the error ratio between sphere test and segment test with L2-Logistic Regression can be found in Figure \ref{fig:logistlabel}. And detailed values can found in Table Table~\ref{Sensitivity2_Log}. We found:

When $C$ increases, the error ratio for both the sphere test and segment test decreases. The difference between the error ratio of the two tests also increases. 
When $C$ is equal to $0.2$ and $P_{up}$ is equal to $10\%$,  the error ratio for the sphere test is almost $7$ times of the error ratio for the segment test.
When $C$ is equal to $0.5$ and $P_{up}$ is equal to $10\%$, the error ratio for the sphere test is almost $8$ times of the error ratio for the segment test.
When $C$ is equal to $1$ and $P_{up}$ is equal to $10\%$, the error ratio for the sphere test is almost $9$ times of the error ratio for the segment test.

This experiment shares the second conclusion of the third experiment.

\begin{table*}[htbp]
	\begin{center}
		\caption{Results for sensitivity of test instance labels task with L2-SVM}\label{Sensitivity2_SVM}
		\scalebox{0.85}{\begin{tabular}{|c|c||c|c|c||c|c|c||c|c|c||c|}
			\hline
			\multicolumn{2}{|c||}{\multirow{2}{*}{}} &
			\multicolumn{10}{|c|}{$C = 0.1$} \\
			\cline{1-12}
			\multicolumn{2}{|c||}{$P_{up}(\%)$}&0.01&0.02&0.05&0.1&0.2&0.5&1&2&5&10\\
			\hline
			\hline
			Sphere Test&Error(\%)&0.04&0.07&0.12&0.20&0.20&0.41&0.65&1.21&2.63&4.88 \\
			\hline
			Segment Test&Error(\%)&0.04&0.07&0.12&0.20&0.20&0.40&0.56&0.91&1.51&2.26 \\
			\hline
			\hline
			\multicolumn{2}{|c||}{\multirow{2}{*}{}} &
			\multicolumn{10}{|c|}{$C = 1$} \\
			\hline
			Sphere Test&Error(\%)&0.02&0.04&0.07&0.12&0.17&0.33&0.59&1.10&2.67&5.10 \\
			\hline
			Segment Test&Error(\%)&0.02&0.04&0.07&0.11&0.16&0.26&0.43&0.57&0.96&1.52 \\
			\hline
			\hline
			\multicolumn{2}{|c||}{\multirow{2}{*}{}} &
			\multicolumn{10}{|c|}{$C = 10$} \\
			\hline
			Sphere Test&Error(\%)&0.02&0.04&0.06&0.09&0.18&0.35&0.62&1.15&2.72&5.25 \\
			\hline
			Segment Test&Error(\%)&0.02&0.04&0.06&0.09&0.17&0.28&0.38&0.48&0.89&1.25 \\
			\hline
		\end{tabular}}
	\end{center}
\end{table*}

\begin{table*}[htbp]
	\begin{center}
		\caption{Results for sensitivity of test instance labels task with L2-Logistic Regression}\label{Sensitivity2_Log}
		\scalebox{0.85}{\begin{tabular}{|c|c||c|c|c||c|c|c||c|c|c||c|}
			\hline
			\multicolumn{2}{|c||}{\multirow{2}{*}{}} &
			\multicolumn{10}{|c|}{$C = 0.1$} \\
			\cline{1-12}
			\multicolumn{2}{|c||}{$P_{up}(\%)$}&0.01&0.02&0.05&0.1&0.2&0.5&1&2&5&10\\
			\hline
			\hline
			Sphere Test&Error(\%)&0.08&0.15&0.26&0.36&0.69&1.65&3.00&5.57&13.73&25.72 \\
			\hline
			Segment Test&Error(\%)&0.01&0.02&0.04&0.06&0.12&0.26&0.45&0.77&1.87&3.44 \\
			\hline
			\hline
			\multicolumn{2}{|c||}{\multirow{2}{*}{}} &
			\multicolumn{10}{|c|}{$C = 1$} \\
			\hline
			Sphere Test&Error(\%)&0.06&0.13&0.20&0.31&0.51&1.27&2.24&4.36&10.40&17.74 \\
			\hline
			Segment Test&Error(\%)&0.01&0.02&0.04&0.06&0.09&0.22&0.32&0.61&1.45&2.22 \\
			\hline
			\hline
			\multicolumn{2}{|c||}{\multirow{2}{*}{}} &
			\multicolumn{10}{|c|}{$C = 10$} \\
			\hline
			Sphere Test&Error(\%)&0.05&0.08&0.16&0.26&0.40&0.96&1.83&3.47&7.72&14.02 \\
			\hline
			Segment Test&Error(\%)&0.01&0.02&0.03&0.06&0.10&0.16&0.28&0.49&0.90&1.52 \\
			\hline
		\end{tabular}}
	\end{center}
\end{table*}

\section{Conclusion}
In this paper, we put forward a sensitivity analysis algorithm named segment test to estimate the bound of a large-scale classifier's parameter matrix without really solving it. The computing complexity of the segment test only depends on the size of the updated dataset which is much lower than the computational complexity of retraining the classifier from scratch. Compare with previous sensitivity analysis algorithms, the segment test can provide a more accurate result without largely increasing the computational complexity and it is more robust when the data set has a relatively huge change. Moreover, the segment test only requires that the estimated linear classifier is calculated based on the differentiable convex L2-regularization loss function, the proposed algorithm can be applied to more diversified cases than conventional incremental learning algorithms which are usually designed for a specific algorithm. 

However, despite these advantages of the segment test, its robustness is still limited when the ratio of modified data set is large and it can only be applied to the linear classifier. Our future research will concentrate on its applicability to the non-linear classifier, e.g., neural network.  

\section{Acknowledgement}
The authors would like to thank Kang Gao (Current PhD student in Imperial College London) for his valuable opinions and useful suggestions.

\bibliography{mybibfile}

\begin{thebibliography}{10}
\expandafter\ifx\csname url\endcsname\relax
  \def\url#1{\texttt{#1}}\fi
\expandafter\ifx\csname urlprefix\endcsname\relax\def\urlprefix{URL }\fi
\expandafter\ifx\csname href\endcsname\relax
  \def\href#1#2{#2} \def\path#1{#1}\fi

\bibitem{zhang2016unsupervised}
X.~Zhang, S.~Song, L.~Zhu, K.~You, C.~Wu, Unsupervised learning of dirichlet
  process mixture models with missing data, Science China Information Sciences
  59~(1) (2016) 1--14.

\bibitem{ristin2016incremental}
M.~Ristin, M.~Guillaumin, J.~Gall, L.~Van~Gool, Incremental learning of random
  forests for large-scale image classification, IEEE transactions on pattern
  analysis and machine intelligence 38~(3) (2016) 490--503.

\bibitem{syed1999incremental}
N.~A. Syed, S.~Huan, L.~Kah, K.~Sung, Incremental learning with support vector
  machines (1999).

\bibitem{cauwenberghs2001incremental}
G.~Cauwenberghs, T.~Poggio, Incremental and decremental support vector machine
  learning, in: Advances in neural information processing systems, Vol.~13,
  2001, pp. 409--415.

\bibitem{Cho2013}
S.~Cho, S.~Jo, {Incremental Online Learning of Robot Behaviors From Selected
  Multiple Kinesthetic Teaching Trials}, IEEE TRANSACTIONS ON SYSTEMS MAN
  CYBERNETICS-SYSTEMS {43}~({3}) ({2013}) {730--740}.

\bibitem{gu2015incremental}
B.~Gu, V.~S. Sheng, K.~Y. Tay, W.~Romano, S.~Li, Incremental support vector
  learning for ordinal regression, IEEE Transactions on Neural networks and
  learning systems 26~(7) (2015) 1403--1416.

\bibitem{karasuyama2009multiple}
M.~Karasuyama, I.~Takeuchi, Multiple incremental decremental learning of
  support vector machines, in: Advances in neural information processing
  systems, 2009, pp. 907--915.

\bibitem{tsai2014incremental}
C.~Tsai, C.~Lin, C.~Lin, Incremental and decremental training for linear
  classification, in: Proceedings of the 20th ACM SIGKDD international
  conference on Knowledge discovery and data mining, ACM, 2014, pp. 343--352.

\bibitem{shilton2005incremental}
A.~Shilton, M.~Palaniswami, D.~Ralph, A.~C. Tsoi, Incremental training of
  support vector machines, IEEE transactions on neural networks 16~(1) (2005)
  114--131.

\bibitem{xu2018new}
J.~Xu, C.~Xu, B.~Zou, Y.~Y. Tang, J.~Peng, X.~You, New incremental learning
  algorithm with support vector machines, IEEE Transactions on Systems, Man,
  and Cybernetics: Systems~(99) (2018) 1--12.

\bibitem{ruping2002incremental}
S.~R{\"u}ping, Incremental learning with support vector machines, Tech. rep.,
  Technical Report, SFB 475: Komplexit{\"a}tsreduktion in Multivariaten
  Datenstrukturen, Universit{\"a}t Dortmund (2002).

\bibitem{ren2010incremental}
C.-X. Ren, D.-Q. Dai, Incremental learning of bidirectional principal
  components for face recognition, Pattern Recognition 43~(1) (2010) 318--330.

\bibitem{saltelli2004sensitivity}
A.~Saltelli, S.~Tarantola, F.~Campolongo, M.~Ratto, Sensitivity analysis in
  practice: a guide to assessing scientific models, John Wiley \& Sons, 2004.

\bibitem{okumura2015kdd}
S.~Okumura, Y.~Suzuki, I.~Takeuchi, Quick sensitivity analysis for incremental
  data modification and its application to leave-one-out cv in linear
  classification problems, in: Proceedings of the 21th ACM SIGKDD International
  Conference on Knowledge Discovery and Data Mining, ACM, 2015, pp. 885--894.

\bibitem{ghaoui2010safe}
L.~E. Ghaoui, V.~Viallon, T.~Rabbani, Safe feature elimination for the lasso
  and sparse supervised learning problems, arXiv preprint arXiv:1009.4219
  (2010).

\bibitem{liu2013safe}
J.~Liu, Z.~Zhao, J.~Wang, J.~Ye, Safe screening with variational inequalities
  and its application to lasso, arXiv preprint arXiv:1307.7577 (2013).

\bibitem{wang2014safe}
J.~Wang, J.~Zhou, J.~Liu, P.~Wonka, J.~Ye, A safe screening rule for sparse
  logistic regression, in: Advances in Neural Information Processing Systems,
  2014, pp. 1053--1061.

\bibitem{xiang2017screening}
Z.~J. Xiang, Y.~Wang, P.~J. Ramadge, Screening tests for lasso problems., IEEE
  Trans. Pattern Anal. Mach. Intell. 39~(5) (2017) 1008--1027.

\bibitem{steinwart2005consistency}
I.~Steinwart, Consistency of support vector machines and other regularized
  kernel classifiers, IEEE Transactions on Information Theory 51~(1) (2005)
  128--142.

\bibitem{lee2013study}
C.-P. Lee, C.-J. Lin, A study on l2-loss (squared hinge-loss) multiclass svm,
  Neural computation 25~(5) (2013) 1302--1323.

\bibitem{zhang2015resting}
X.~Zhang, B.~Hu, X.~Ma, L.~Xu, Resting-state whole-brain functional
  connectivity networks for mci classification using l2-regularized logistic
  regression, IEEE transactions on nanobioscience 14~(2) (2015) 237--247.

\bibitem{cox1958regression}
D.~R. Cox, The regression analysis of binary sequences, Journal of the Royal
  Statistical Society. Series B (Methodological) (1958) 215--242.

\bibitem{xiang2014screening}
Z.~Xiang, Y.~Wang, P.~Ramadge, Screening tests for lasso problems, arXiv
  preprint arXiv:1405.4897 (2014).

\bibitem{lorensen1987marching}
W.~E. Lorensen, H.~E. Cline, Marching cubes: A high resolution 3d surface
  construction algorithm, in: ACM siggraph computer graphics, Vol.~21, ACM,
  1987, pp. 163--169.

\bibitem{maurer1979first}
H.~Maurer, J.~Zowe, First and second-order necessary and sufficient optimality
  conditions for infinite-dimensional programming problems, Mathematical
  programming 16~(1) (1979) 98--110.

\bibitem{hummel1949generalization}
P.~Hummel, C.~Seebeck~Jr, A generalization of taylor's expansion, The American
  Mathematical Monthly 56~(4) (1949) 243--247.

\bibitem{boyer1994influence}
L.~Boyer, F.~Houze, A.~Tonck, J.-L. Loubet, J.-M. Georges, The influence of
  surface roughness on the capacitance between a sphere and a plane, Journal of
  Physics D: Applied Physics 27~(7) (1994) 1504.

\bibitem{borneas1959generalization}
M.~Borneas, On a generalization of the lagrange function, American Journal of
  Physics 27~(4) (1959) 265--267.

\bibitem{wu2007karush}
H.-C. Wu, The karush--kuhn--tucker optimality conditions in an optimization
  problem with interval-valued objective function, European Journal of
  Operational Research 176~(1) (2007) 46--59.

\bibitem{chang2011libsvm}
C.~Chang, C.~Lin, Libsvm: a library for support vector machines, ACM
  Transactions on Intelligent Systems and Technology (TIST) 2~(3) (2011) 27.

\bibitem{chen1998linear}
C.-T. Chen, Linear system theory and design, Oxford University Press, Inc.,
  1998.

\end{thebibliography}
\par\noindent 
\parbox[t]{\linewidth}{
\noindent\parpic{\includegraphics[height=1.5in,width=1in,clip,keepaspectratio]{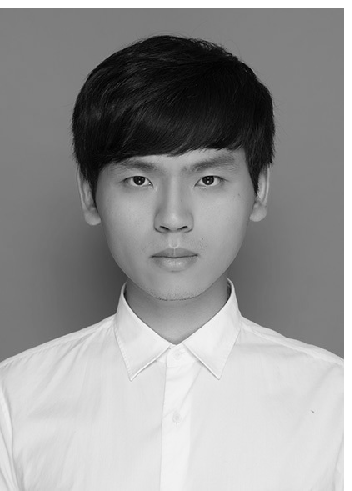}}
\noindent {\bf Kaichen Zhou}\
received the Master degree in Machine Learning in the Department of Computing at Imperial College London. He is currently pursuing the P.h.D degree in Computer Science in the Department of Computer Science at University of Oxford. His research interests include deep learning and reinforcement learning. }
\vspace{4\baselineskip}

\par\noindent 
\parbox[t]{\linewidth}{
\noindent\parpic{\includegraphics[height=1.5in,width=1in,clip,keepaspectratio]{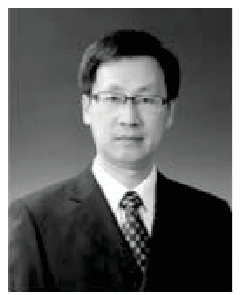}}
\noindent {\bf Shiji Song}\
received the Ph.D. degree from the Department of Mathematics, Harbin Institute of Technology, Harbin, China, in 1996. He is currently a Professor with the Department of Automation, Tsinghua University, Beijing, China. His current research interests include system modeling, control and optimization, computational intelligence, and pattern recognition.}
\vspace{4\baselineskip}

\par\noindent 
\parbox[t]{\linewidth}{
\noindent\parpic{\includegraphics[height=1.5in,width=1in,clip,keepaspectratio]{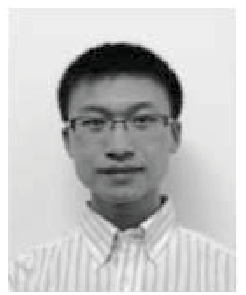}}
\noindent {\bf Gao Huang}\
received the B.S. degree from the School of Automation Science and Electrical Engineering,	Beihang University, Beijing, China, in 2009, and the Ph.D. degree from the Department of	Automation, Tsinghua University, Beijing, in 2015. He was a Visiting Research Scholar with the Department of Computer Science and Engineering, Washington University in St. Louis, St. Louis, MO, USA, in 2013. He is currently a Assistant Professor with the Department of Automation, Tsinghua University, Beijing, China. His current research interests include machine learning and statistical pattern recognition.}
\vspace{4\baselineskip}

\par\noindent 
\parbox[t]{\linewidth}{
\noindent\parpic{\includegraphics[height=1.5in,width=1in,clip,keepaspectratio]{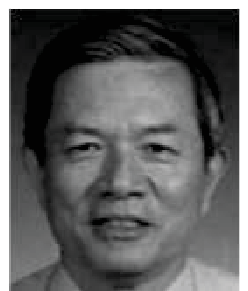}}
\noindent {\bf Cheng Wu}\
received the B.S. and M.S. degrees in electrical engineering from Tsinghua University, Beijing, China. Since 1967, he has been with Tsinghua University, where he is currently a Professor with the Department of Automation. His current research interests include system integration, modeling, scheduling, and optimization of complex industrial systems. Mr. Wu is a member of the Chinese Academy of Engineering.}
\vspace{4\baselineskip}

\par\noindent 
\parbox[t]{\linewidth}{
\noindent\parpic{\includegraphics[height=1.5in,width=1in,clip,keepaspectratio]{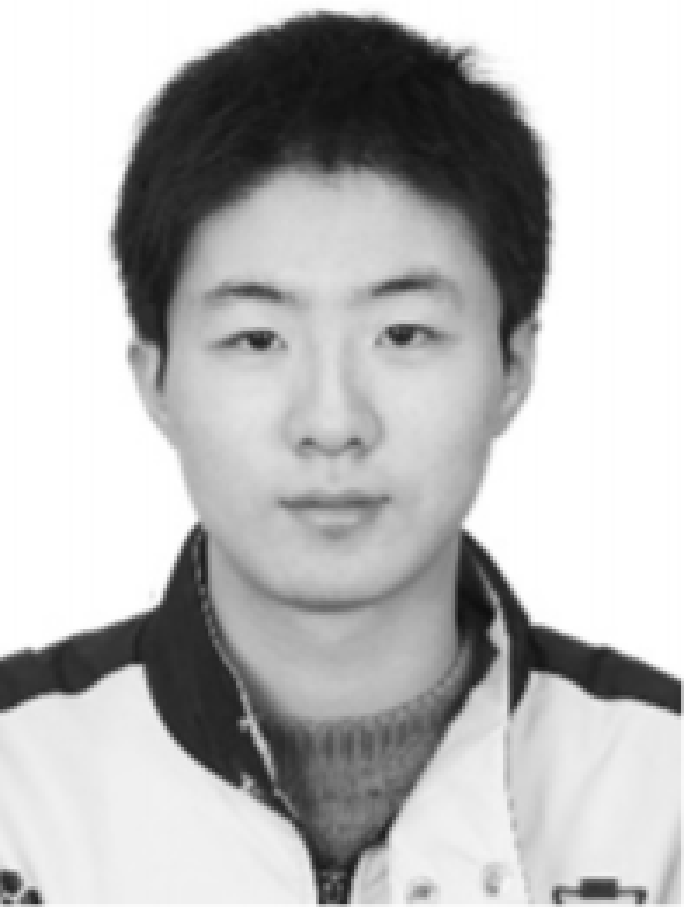}}
\noindent {\bf Quan Zhou}\
received the B.S. degree from the School of Automation
Science and Electrical Engineering, Beihang University, Beijing, China. He was a Visiting Research Scholar with the Department of Computer Science and Engineering, Washington University in St. Louis, St. Louis, MO, USA, in 2014. His current research interests include machine learning and statistical learning.}
\vspace{4\baselineskip}

\end{document}